\documentclass[11pt,a4paper]{article}
\usepackage[hyperref]{eacl2021}
\usepackage{times}
\usepackage{latexsym}

\usepackage{ textcomp }

\usepackage{pifont}

\newcommand{\comment}[1]{}

\usepackage{multirow, tabu}
\usepackage{hyperref}

\usepackage{microtype}

\usepackage{microtype}
\usepackage{amsmath}
\usepackage{amssymb}
\usepackage{bbm}
\usepackage{graphicx}
\usepackage{multicol}
\usepackage{tabularx}
\usepackage{booktabs}
\usepackage{algorithmic}
\usepackage{algorithm}
\usepackage{cleveref}
\usepackage{multirow}

\usepackage{xspace} 

\newcommand{\wa}{\textit{weak annotator}\xspace}
\usepackage{bm}

\newcommand{\xmark}{\ding{55}}%

\newcommand{\std}{student\xspace}
\newcommand{\tch}{teacher\xspace}

\DeclareFontFamily{U}{mathb}{\hyphenchar\font45}
\DeclareFontShape{U}{mathb}{m}{n}{
<-6> mathb5 <6-7> mathb6 <7-8> mathb7
<8-9> mathb8 <9-10> mathb9
<10-12> mathb10 <12-> mathb12
}{}
\DeclareSymbolFont{mathb}{U}{mathb}{m}{n}

\usepackage{ wasysym }

\crefformat{section}{\S#2#1#3}
\crefformat{subsection}{\S#2#1#3}
\crefformat{subsubsection}{\S#2#1#3}

\definecolor{darkgreen}{RGB}{84,174,50}

\aclfinalcopy 
\def\aclpaperid{332} 

\title{Jointly Improving Language Understanding and Generation with Quality-Weighted Weak Supervision of Automatic Labeling}

\author{ 
  \dag Ernie Chang,  \dag Vera Demberg, *Alex Marin \\
  \dag Dept. of Language Science and Technology, Saarland University \\
    {\tt \{cychang,vera\}@coli.uni-saarland.de}
  \\ *Microsoft Corporation, Redmond, WA
  \\ 
  {\tt \{alemari\}@microsoft.com}\\
}

\date{}

\begin{document}
\maketitle
\begin{abstract}
Neural natural language generation (NLG) and understanding (NLU) models are data-hungry and require massive amounts of annotated data to be competitive.
Recent frameworks address this bottleneck with generative models that synthesize weak labels at scale, where a small amount of training labels are expert-curated and the rest of the data is automatically annotated.
We follow that approach, by automatically constructing a large-scale \emph{weakly-labeled data} with a fine-tuned GPT-2,
and employ a semi-supervised framework
to jointly train the NLG and NLU models. 
The proposed framework adapts the parameter updates to the models according to the estimated label-quality.
On both the E2E and Weather benchmarks, we show that this weakly supervised training paradigm is an effective approach under low resource scenarios with as little as $10$ data instances, and outperforming benchmark systems on both datasets when $100$\% of training data is used. 
\end{abstract}

\section{Introduction}
Natural language generation (NLG) is the task that transforms meaning representations (MR) into natural language descriptions~\cite{reiter2000building,barzilay-lapata-2005-modeling};
while natural language understanding (NLU) is the opposite process where text is converted into MR~\cite{zhang2016joint}.
These two processes can thus constrain each other -- recent exploration of the duality of neural natural language generation (NLG) and understanding (NLU) has led to successful semi-supervised learning techniques where both labeled and unlabeled data can be used for training~\cite{su2020towards,tseng2020generative,schmitt2019unsupervised,qader2019semi,su2020towards}. 


Standard supervised learning for NLG and NLU depends on the access to labeled training data -- a major bottleneck in developing new applications. 
In particular, neural methods require a large annotated dataset for each specific task. 
The collection process is often prohibitively expensive, especially when specialized domain expertise is required. 
\begin{figure}
  \centering
\includegraphics[width=0.95\columnwidth]{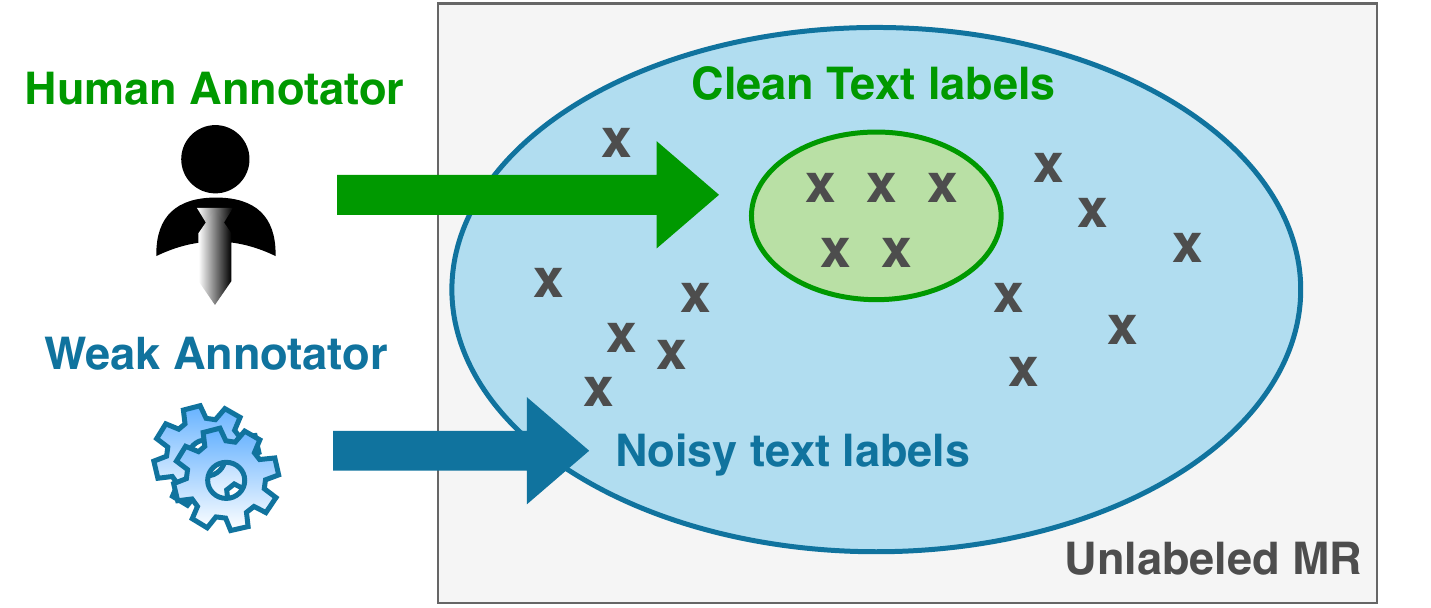}
\caption{ \small
\textbf{Training scenario}: Each \textbf{\texttimes} represents a labeled data instance. The goal is to learn both from few human-labeled instances (inner) and large amounts of weakly labeled data (outer). }
\label{fig:sample}
\end{figure}
\comment{
Further, past semi-supervised learning approaches \emph{assume the presence of vast amounts of unpaired MR and text instances}. However, this is often not the case: while MR samples are often abundant, text samples are usually harder to come by (and are often paired with MRs) \cite{chu2019meansum}. 
Constructing sentences for NLG requires adhering strictly to a predefined MR schema. Obtaining high-quality, even if unpaired, text is thus challenging in itself.}
On the other hand, learning with weak supervision from noisy labels offers a potential solution as it automatically builds imperfect training sets from low cost labeling rules or pretrained models~\cite{zhou2018brief,ratner2017snorkel,fries2020trove}. 
Further, labeled data and large unlabeled data can be utilized in semi-supervised learning~\cite{lample2017unsupervised,tseng2020generative}, as a way to \emph{jointly} improve both NLU and NLG models. 

To this end, we target a weak supervision scenario (shown in Figure~\ref{fig:sample}) consisting of small, high-quality expert-labeled data and a large set of unlabeled MR instances. 
We propose to expand the labeled data by automatically annotating the MR samples with noisy text labels. 
These noisy text labels are generated by a \wa, which is built upon recent works that directly fine-tune GPT-2~\cite{GPT2} on joint meaning representation (MR) and text~\cite{mager2020gpt,harkous2020have}. 
Then, we jointly train the NLG and NLU models in a two-step process with semi-supervised learning objectives~\cite{tseng2020generative}. First, we use pretrained models to estimate quality scores for each sample. Then, we down-weight the loss updates in the back-propagation phase using the estimated quality scores. This way, the models are guided to avoid mistakes of the \wa.

\begin{figure*}[ht]
  \centering
\includegraphics[width=1\textwidth]{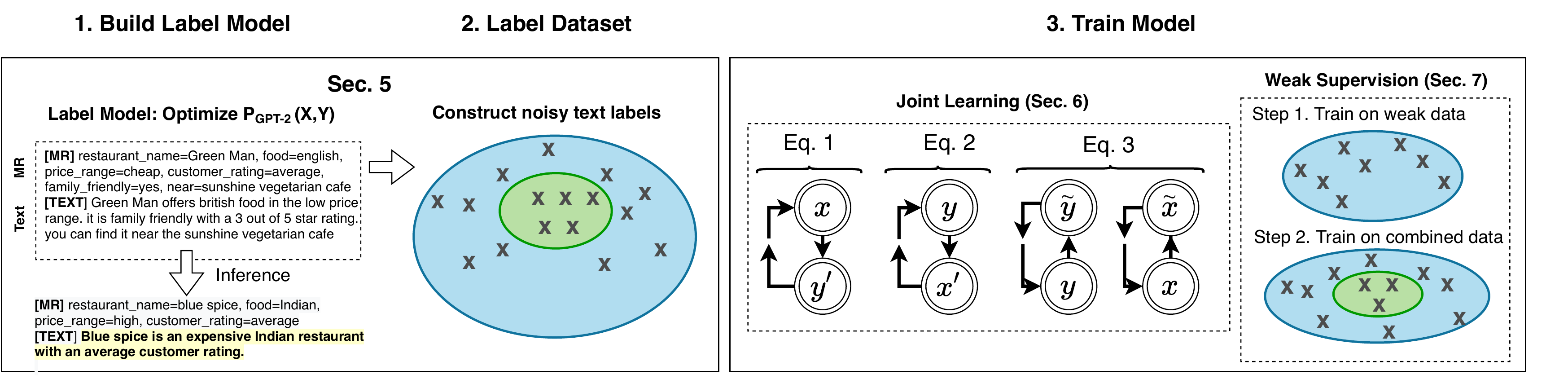}
\caption{\small Depiction of the proposed framework.
In \emph{joint learning}, gradients are back-propagated through solid lines.
}
\label{fig:sum}
\end{figure*}

On two benchmarks,  E2E~\cite{novikova2017e2e} and Weather~\cite{balakrishnan2019constrained}, we utilize varying amount of labeled data and show that the framework is able to successfully learn from the synthetic data generated by \emph{weak annotator}, thereby allowing jointly-trained NLG and NLU models to outperform other baseline systems.

This work makes the following contributions:
\begin{enumerate}
    \item We propose an automatic method to overcome the lack of text labels by using a fine-tuned language model as a \emph{weak annotator} to  construct text labels for the vast amount of MR samples, resulting in a much larger labeled dataset. 
    \item We propose an effective two-step weak supervision using the dual mutual information (DMI) measure which can be used to modulate parameter updates on the weakly labeled data by providing quality estimates.
    \item We show that the approach can even be used to improve upon baselines with 100\% data to establish new state-of-the-art performance.
\end{enumerate}


\section{Related Work}
\paragraph{Learning with Weak Supervision.}
Learning with weak supervision is a well-studied area that is popularized by the rise of data-driven neural approaches~\cite{ratner2017snorkel,safranchik2020weakly,bach2017learning,wu2018learning,dehghani2018fidelity,jiang2018mentornet,chang2020unsupervised, de2018generating}.
Our approach incorporates similar line of work, by providing noisy labels (text) with a fine-tuned LM which incorporates prior knowledge from general-domain text and data-text pair~\cite{budzianowski2019hello,chen2019few, peng2020few, mager2020gpt, harkous2020have, shen2020neural, chang2020dart, chang2021does, chang2021neural}, and use it as the \emph{weak annotator}, similar by functionality to that of fidelity-weighted learning~\cite{dehghani2017fidelity}, or data creation tool \emph{Snorkel}~\cite{ratner2017snorkel}.

\paragraph{Learning with Semi-Supervision.}
Work on semi-supervised learning considers settings with some labeled data and a much larger set of unlabeled data, and then leverages both labeled the unlabeled data as in machine translation~\cite{artetxe2017unsupervised,lample2017unsupervised}, data-to-text generation~\cite{schmitt2019unsupervised,qader2019semi} or more relevantly the joint learning framework for training NLU and NLG~\cite{tseng2020generative,su2020towards}. 
Nonetheless, these approaches all assume that a \emph{large collection of text} is available, which is an unrealistic assumption for the task due to the need for expert curation.
In our work, we show that both NLU and NLG models can benefit from (1) automatically labeling MR with text, and (2) by semi-supervisedly learning from these samples while accounting for their qualities.

\section{Approach}
\label{sec:length}
We represent the set of meaning representation (MR) as \textit{X} and the text samples as \textit{Y}. There are no restrictions on the format of the MR: each $x \in X$ can be a set of slot-value pairs, or can take the form of tree-structured semantic definitions as in \citet{balakrishnan2019constrained}. Each text $y \in Y$ consists of a sequence of words.

In our setting, we have (1) $k$ labeled pairs and (2) a large quantity of unlabeled MR set $X_U$ where $|X_U| \gg k > 0$. (We force $k>0$ as we believe a reasonable generation system needs at least a few demonstrations of the annotation.) 
This is a realistic setting for novel application domains, 
as unlabeled MR are usually abundant and can also be easily constructed from predefined schemata. 
Notably, \emph{we assume no access to outside resources containing in-domain text}. 
The $k$ annotations are all we know about in-domain text.

The core of our approach consists of first labeling MR samples with text, and then training on the expanded dataset.
We start with describing the process of creating weakly labeled data~(\cref{sec:aug}).
Next, we delve into the semi-supervised training objectives for the NLU and NLG models, which allow the models to learn from labeled and unlabeled data~(\cref{sec: back-and-forth}).
Lastly, we explain the training process where NLG and NLU models are jointly optimized in two steps: 
In \emph{step 1}, we pretrain the models on the weakly-labeled corpus, then \emph{continue updating the models} on the combined data consisting of the weak and real data in \emph{step 2}.
Importantly, to account for the noise that comes with the automatic weak annotation, \emph{step 2} trains the model with quality-weighted updates~(\cref{sec: rm}).
We depict this process in Figure~\ref{fig:sum}.

\section{ Creating Weakly Labeled Data}
\label{sec:aug}

We construct synthetic data in two ways:
(1) creating more MR samples (see~\cref{sec:mr}), and
(2) by creating a larger parallel set of MRs with texts (see~\cref{sec:text-aug}). 

\subsection{Generating Synthetic MR Samples}
\label{sec:mr}
We consider a simple way of MR
augmentation via value swapping. 
This creates more unlabeled MR to be annotated by the \emph{weak annotator} and also provide a substantial augmentation that benefits the autoencoding on MR samples (see Equation~\ref{eq: ae}) by exposing it to a larger set of MR.
\begin{figure}[H]
  \centering
\includegraphics[width=0.8\columnwidth]{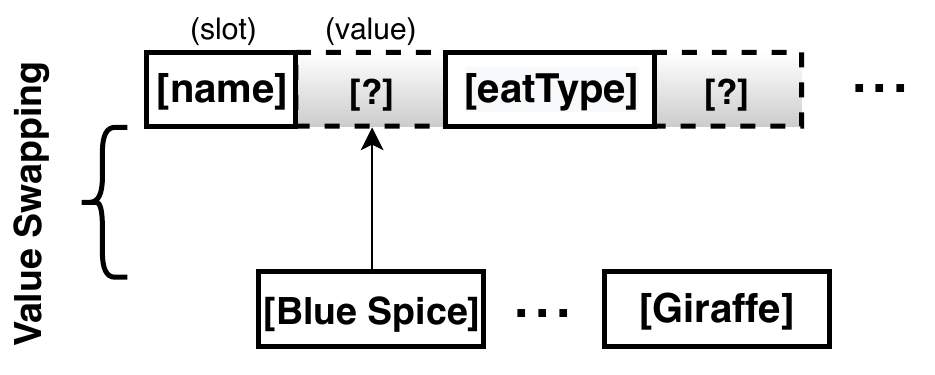}
\caption{\small Depiction of MR augmentation in the E2E corpus.}
\label{fig:swap}
\end{figure}
Since each slot in the MR samples corresponds to multiple possible 
values, we pair each \emph{slot} with a randomly sampled value collected from the set of all MR samples to obtain new combination of slot-value pairs. 
This way, we create a large synthetic MR set.

\subsection{Creation of Parallel MR-to-Text Set}
\label{sec:text-aug}
GPT-2~\cite{GPT2} is a powerful language model pretrained on the large WebText corpus. 
Recent work
on conditional data-to-text generation~\cite{harkous2020have,mager2020gpt} demonstrated that fine-tuning GPT-2 on the joint distribution of MR and text for text-only generation yields impressive performance. 

The fine-tuned model generates in-domain text by conditioning on samples from the augmented MR set ($X_{U}$).
Rather than using GPT-2 outputs directly, we employ them in a process analogous to knowledge distillation~\cite{tan2018multilingual,tang2019natural,baziotis2020language} where the fine-tuned GPT-2 provides supervisory signals instead of being used directly for generation.

We now describe the process of GPT-2 fine-tuning.
Given the sequential MR representation $x_1 \cdots x_M$ and a sentence $y_1 \cdots y_N$ in the labeled dataset $(X_{L},Y_{L})$, we maximize the joint probability $p_{\mbox{\scriptsize GPT-2}}(X_{L},Y_{L})$, 
where each sequence is concatenated into ``[MR] $x_1 \cdots x_M$ [TEXT] $y_1 \cdots y_N$''.  
In addition, we also freeze the input embeddings when fine-tuning had positive impact on performance, following ~\citet{mager2020gpt}. 
At test time, we provide the MR samples as context as in conventional conditional text generation:

\begin{equation}
\tilde{y}_j = \arg\max_{y_j} \{p_{\mbox{\scriptsize GPT-2}}(y_j \mid y_{1:j-1}, x_{1:N}) \}\nonumber
\end{equation}

 The fine-tuned LM conditions on augmented MR sample set $X_{U}$ to generate the in-domain text\footnote{We adopt the Top-$k$ random sampling with $k=2$ to encourage diversity and reduce repetition~\cite{GPT2}}, forming the weak label dataset $\mathcal{D}_W=(X_{U}, \tilde{Y}_{L})$ with noisy labels $\tilde{y}_i \in \tilde{Y}_{L}$. 
In practice, the fine-tuned LM produces malformed, synthetic text which does not fully match with the MR it was conditioned on, 
as it might hallucinate additional values not consistent with its MR counterpart.
Thus, it is necessary to check for factual consistency~\cite{moryossef2019step}. We address this point next.

Past findings showed
(e.g.~\cite{wang2019revisiting}) that the removal of utterance with ``hallucinated'' facts (MR values) from MR
leads to considerable performance gain, since inconsistent MR-Text correspondence might misguide systems to generate incorrect facts and deteriorate the NLG outputs. 
We filter out the synthetic, poor quality MR-text pairs by
training a separate NLU model on the original labeled data to predict MR from generated text labels. 
These MRs can then be checked against the paired MR in $\mathcal{D}_W$ via pattern matching as inspired by~\citet{cai2013smatch,wiseman-etal-2017-challenges}.
Specifically, we use a measure of semantic similarity in terms of \emph{f-score via matching of slots between the two MRs}.
We keep all MR-text pairs with f-scores above $0.7$, as we found empirically that this criterion retains a sufficiently large amount of high-quality data.
The removed text sentences
are used for unsupervised training objectives as in Eq.~\ref{eq: dtd}-\ref{eq: ae}.
Using this method, we create a collection of parallel MR-text samples (\texttildelow
 500k) an order of magnitude
 larger than even the full training sets (\texttildelow 40k for E2E and \texttildelow 25k for Weather).

\section{Joint learning of NLG and NLU} 
\label{sec: back-and-forth}
For both NLU and NLG models, we adopt the same architecture as~\citet{tseng2020generative}, which use two Bi-LSTM-based~\cite{hochreiter1997long} encoders for each model. 
The NLU decoder for slot-value structured data \citep[e.g., E2E,][]{mrksic-etal-2017-neural}
contains several 1-layer feed-forward neural classifiers for each slot;
while for tree-structured meaning representation in~\citet{balakrishnan2019constrained}, the decoder is LSTM-based.
\comment{
\begin{algorithm}[b]
   \caption{Training Process}
   \label{alg}
\begin{algorithmic}[1]
   \STATE {\bfseries Input:} $X_U, Y_U, D $
   \comment{\STATE Create $(X_a, Y_a)$ by MR augmentation (\cref{sec: mr-aug});
   \STATE $(X_L, Y_L) \leftarrow (X_L, Y_L) \cup (X_a, Y_a)$;
   \STATE $Y \leftarrow Y_L \cup Y'$;
   \STATE $Y \leftarrow Y_L \cup Y'$;}
   \STATE \textbf{Unsupervised Learning on $\{ X_U, Y_U \}$:}
   \STATE Optimize by Eq.~\ref{eq: dtd} + Eq.~\ref{eq: tdt} + Eq.~\ref{eq: ae};   
   \STATE \textbf{Supervised Learning on $ \{ \mathcal{D} \}$:}
   \STATE Optimize by Eq.~\ref{eq: sup};   
\end{algorithmic}
\end{algorithm}}
In this framework, both NLU and NLG models are trained to infer the shared latent variable repeatedly -- starting from either MR or text, in order to encourage semantic consistency.
Each model can be improved via gradient passing between them using REINFORCE~\cite{williams1992simple}. 
This way, the models benefit from each other's training in a process known as the \emph{dual learning}~\cite{su2020towards}, which consists of both \emph{unsupervised} and \emph{supervised} learning objectives.
We now go into details describing them.

\paragraph{Unsupervised Learning.} Starting from either a MR sample or a text sample,
the models project the sample from one space into the other, then map it back to the original space (either MR or text sample, respectively), and compute the reconstruction loss after the two operations.
This repetition will result in aligned pairs between the MR samples and corresponding text~\cite{he2016dual}. 
Specifically, let $p_\theta(y|x)$ be the probability distribution to map $x$ to its corresponding $y$ (NLG), and $p_\phi(x|y)$ be the probability distribution to map $y$ back to $x$ (NLU). 
Starting from $x \in X$, its objective is:
\begin{equation}
\label{eq: dtd}
\begin{split}
    \max_{\phi}      \mathbb{E}_{x \sim p(X)}\log p_\phi(x|y');\;
    y' \sim p_\theta(y|x)
\end{split}
\end{equation}
which ensures the semantic consistency by first performing NLG accompanied by NLU in direction $x \rightarrow y' \rightarrow x$. 
Note that only $p_\phi$ is updated in this direction and $p_\theta$ serves only as as an auxiliary function to provide pseudo samples $y'$ from $x$. 
\comment{Similar observations are also found in \citet{lample2018phrase,he2020probabilistic,garcia2020multilingual}.}
\comment{
\begin{figure}[t!]
  \centering
\includegraphics[width=\columnwidth]{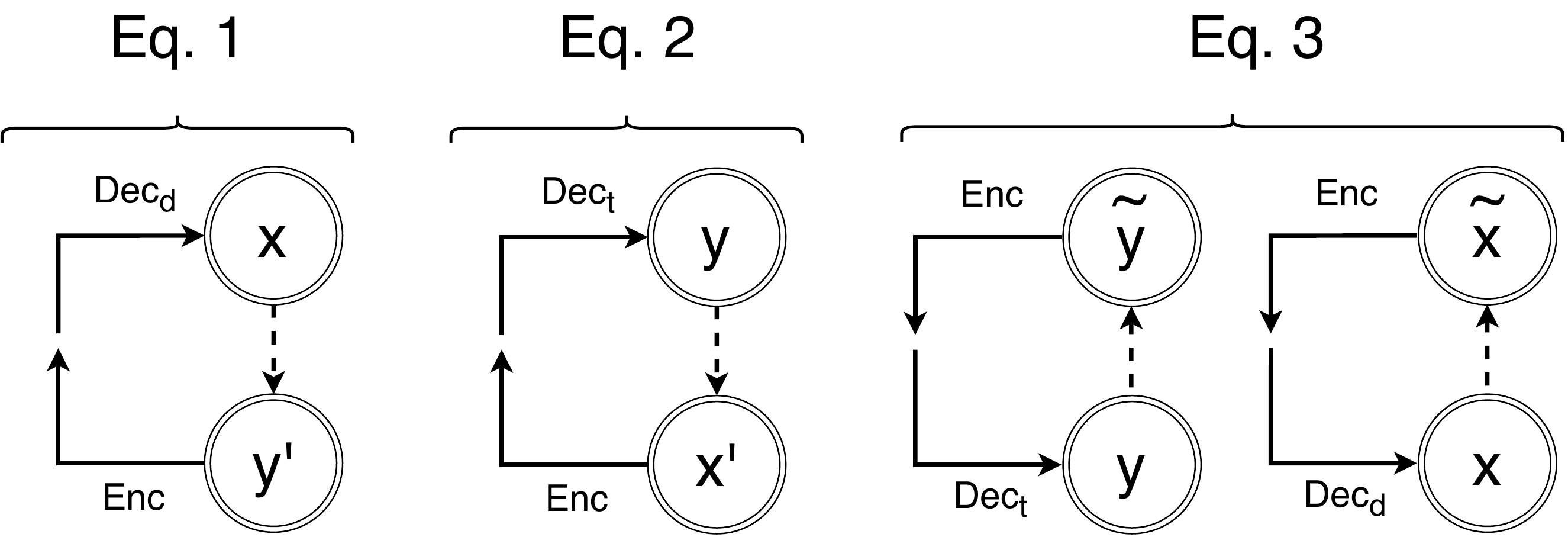}
\caption{\small Four directions of dual learning. Gradients are back-propagated only through solid lines.}
\label{fig:directions}
\end{figure}}
Similarly, starting from $y \in Y$, the objective ensures semantic consistency in the direction where the NLU step is followed by NLG: $y \rightarrow x' \rightarrow y$\footnote{This direction is usually termed as \emph{back translation} in MT community~\cite{sennrich2016improving,lample2018phrase}}:
\begin{equation}
\label{eq: tdt}
\begin{split}
    \max_{\theta}      \mathbb{E}_{y \sim p(Y)}\log p_\theta(y|x');\;x' \sim p_\phi(x|y)
\end{split}
\end{equation}
We further add two autoencoding objectives on both MR and text samples:
\begin{equation}
\label{eq: ae}
    \max_{\theta,\phi} \mathbb{E}_{x \sim p(X), y \sim p(Y)} \log p_\phi(x|{x})p_\theta(y|{y}) 
\end{equation}
Thus, unlabeled text samples can be used as they are shown to benefit the text space ($Y$) by introducing new signals into learning directions $y \rightarrow x' \rightarrow y$ and $\Tilde{y}\rightarrow y$.
Thus, \emph{we use all in-domain text data} whether they have corresponding MR or not. 
Note that following~\cite{tseng2020generative}, we also adopt the variational optimization objective upon the latent variable $z$ which was shown to pull the inferred posteriors $q(z|x)$ and $q(z|y)$ closer to each other. In this case, the parameters of both NLG and NLU models are updated.

\paragraph{Supervised Learning.} Apart from the above unsupervised objectives, we can impose the supervised objective on the $k$ labeled pairs:
\begin{equation}
\label{eq: sup}
    \max_{\theta,\phi} \mathbb{E}_{x,y \sim p(X_L,Y_L)} \log p_\theta(y|x) + \log p_\phi (x|y)
\end{equation}
Each MR is flattened into a sequence and fed into the NLG encoder, giving NLG and NLU models an inductive bias to project similar MR/text into the surrounding latent space~\cite{chisholm2017learning}. 
As we observed anecdotally\footnote{\citet{tseng2020generative} noticed similar trend in the experiments.}, the information flow enabled by REINFORCE allows the models to utilize unlabeled MR and text, boosting the performance in our scenarios. 
\comment{From the shared encoded space, decoder is utilized to decode $d$ and $t$ respectively.} 

\section{Learning with Weak Supervision}
\label{sec: rm}
The primary challenge that arises from the synthetic data is the \emph{noise} introduced during the generation process.
Noisy and poor quality labels tend to bring little to no improvements~\cite{elman1993learning,frenay2013classification}.
To better train on the large and noisy corpus described in section~\cref{sec:aug} (size \texttildelow
 500k), we employ a two-step training process motivated by fidelity-weighted learning~\cite{dehghani2018fidelity}. 
 The two-step process consists of (1) \emph{pretraining} and (2) \emph{quality-weighted fine-tuning} to account for the heterogenous data quality.

\paragraph{Step 1: \emph{Pre-train two sets of models on weak and clean data, respectively. }} 

We train the first set of models (\emph{teacher}) consisting of NLU, NLG, and autoencoder (AUTO) models on the clean data.
The second set of models (i.e. NLU and NLG) is the \emph{student} that  pretrains on the weak data.


\vspace{-2pt}
\paragraph{Step 2: \emph{Fine-tune the student model parameters on the \textbf{combined clean and weak datasets}.}}

We use each \emph{teacher model} to determine the step size for each iteration of the stochastic gradient descent (SGD) by down-weighting the training step of the corresponding \emph{student model} using the sample quality given by the teacher.
Data points with true labels will have high quality, and thus will be given a larger step-size when  updating the parameters; conversely,  
we down-weight the training steps of the \std for data points where the \tch is not confident.
For this specific fine-tuning process, we update the parameters of the \std (i.e. NLG and NLU models) at time $t$ by training with SGD, 
\comment{where $l(\cdot)$ is the per-example loss\footnote{In practice, we average the computed $\eta_t$ over the batch, while keeping the batch-size small.}, $\eta_t$ is the total learning rate, $N$ is the size of the dataset $\mathcal{D}_{sw}$, $\pmb{w}$ is the parameters of the \std model.}
where $\mathcal{L}(\cdot)$ is the loss of predicting $\hat{y}$ for an input $x_i$ when the label is $\tilde{y}$. 
The weighted step is then $c(x_i, \tilde{y}_i)  \nabla \mathcal{L}( \hat{y}, \tilde{y})$, where $c(\cdot)$ is a scoring function learned by the \emph{teacher} taking as input MR $x_i$ and its noisy text label $\tilde{y}_i$. 
In essence, we control the degree of parameter updates to the \emph{student} based on how reliable its labels are according to the \emph{teacher}.

We denote $c(\cdot)$ as the function of the label quality based on the \emph{dual mutual information  ($\text{DMI}$)}, defined as the \emph{absolute} difference between mutual information ($\text{MI}$)\footnote{Mutual information for $x \rightarrow y$ can be seen as $H(x \rightarrow y) = H_{AUTO}(y) - H_{NLG}(y|x)$~\cite{bugliarello2020s}. } in inference directions $x \rightarrow y$ and $y \rightarrow x$.
\citet{bugliarello2020s} shows that $\text{MI}_{x \rightarrow y}$ correlates to the difficulty in predicting $y$ from $x$, and vice versa. 
Thus we expect the difference between $\text{MI}_{x \rightarrow y}$ and $\text{MI}_{y \rightarrow x}$ for clean sample $(x,y)$ to be relatively small compared to noisy samples, since the level of difficulty is largely \emph{proportional} between NLU and NLG on the samples -- difficulty in inferring $x$ from $y$ will result in harder prediction of $y$ from $x$.
Based on this intuition, the $\text{DMI}$ score of the sample $(x,y)$ is defined as:
\begin{equation}
exp \bigg\{ \left\lvert  \underbrace{\log \frac{q_{\text{AUTO}}(y)}{q_{\text{NLG}}(y | x)}}_{\text{MI}_{x \rightarrow y}} -
\underbrace{\log \frac{q_{\text{AUTO}}(x)}{q_{\text{NLU}}(x | y)}}_{\text{MI}_{y \rightarrow x}}
  \right\rvert \bigg\}. \nonumber
\end{equation}
where $q(\cdot)$ are the two respective models.
The DMI for a clean MR-text pair should be \emph{relatively small}, as the two sides contain proportional semantic information\footnote{We found that mutual information for $x \rightarrow y$ is usually greater than that of $y \rightarrow x$ since NLG is a one-to-many and more difficult process as opposed to NLU.}, and so poor quality samples tend to have higher DMI scores and lower $c(\cdot)$ as they are \emph{less semantically aligned}.
Thus, $c(\cdot)$ defines the \emph{confidence} (quality) the teacher has about the current MR-text sample. We use $c(\cdot)$ to scale $\eta_t$. 
Note that $\eta_t(t)$ does not necessarily depend on each data point, whereas $c(\cdot)$ does. 
We define $c(x_t,y_t)$ as:
\begin{equation}
 \label{eqn:eta2}
 c(x_t,y_t) = 1-\mathcal{N}(\text{DMI}(x_t,y_t)) \nonumber
\end{equation}
where $\mathcal{N}(\cdot)$ normalizes DMI over all samples in both clean and weak data to be in $[0,1]$.
\comment{
to exponentially decrease the learning rate for data point $x_t$ if its corresponding soft label $y_t$ is unreliable (far from a true sample). 
In Equation~\ref{eqn:eta2}, $\beta$ is a positive scalar hyper-parameter. 
Intuitively, small $\beta$ results in a \std which listens more carefully to the \tch and copies its knowledge, while a large $\beta$ makes the \std pay less attention to the \tch, staying with its initial weak knowledge. 
More concretely speaking, as $\beta \to 0$ \std places more trust in the labels $y_t$ estimated by the \tch and the \std copies the knowledge of the \tch. 
On the other hand, as $\beta \to \infty$, \std puts less weight on the extrapolation ability of $\mathcal{GP}$ and the parameters of the \std are not affected by the correcting information from the \tch. }

\begin{table*}
  \small
  \centering
  \resizebox{0.8\textwidth}{!}{
  \begin{tabular}{lcccccccccc}
    \toprule
    \multirow{2}{*}{Model} & \multicolumn{5}{c}{E2E (NLG)} & \multicolumn{5}{c}{E2E (NLU)} \\
    \cmidrule(r){2-6}
    \cmidrule(r){7-11}
     & 10 & 50 & 1\% & 5\% & 50\% & 10 & 50 & 1\% & 5\% & 50\% \\
    \cmidrule(r){1-1}
    \cmidrule(r){2-6}
    \cmidrule(r){7-11}
    WA
    & 0.195
    & 0.287
    & 0.563
    & 0.649
    & 0.714
    & 9.48
    & 11.66
    & 13.20
    & 45.21
    & 65.81 \\  
    $\textup{JUG}^{*}$
     & 0.002  & 0.015  & 0.726 & 0.7671 & 0.819
     & 0.00  & 0.00  & 32.24  & 53.20 & \textbf{78.93 } \\
    \cmidrule(r){1-1}
    \cmidrule(r){2-6}
    \cmidrule(r){7-11}     
    decoupled & 0.261 & 0.279 & 0.648 & 0.693 & 0.793 & 0.00  & 0.00  & 20.51  & 52.77  & 73.68  \\
    joint 
    & 0.218
    & 0.336
    & 0.732
    & 0.764
    & 0.775
    & 0.00   & 6.18   & 24.98   & 49.66  & 70.33 \\    
    joint+aug
    & 0.275
    & 0.381
    & 0.748
    & 0.781
    & 0.797
    & 5.88   & 15.79 & 25.15  & 53.20  & 69.68 \\
    step 1
    & 0.441
    & 0.487
    & 0.610
    & 0.642
    & 0.685
    & 13.18 & 14.28 & 15.37 & 44.72  & 65.20 \\ 
    Ours (step 1+2)
    & \textbf{0.489}
    & \textbf{0.558}
    & \textbf{0.754}
    & \textbf{0.775}
    & \textbf{0.822}
    & \textbf{15.81 }
    & \textbf{23.67 }
    & \textbf{34.09 }
    & \textbf{56.33 }
    & 72.45 \\ 
    \midrule
    \midrule& \multicolumn{5}{c}{Weather (NLG)} & \multicolumn{5}{c}{Weather (NLU)} \\
    \cmidrule(r){2-6}
    \cmidrule(r){7-11}
    Model & 10 & 50 & 1\% & 5\% & 50\% & 10 & 50 & 1\% & 5\% & 50\% \\
    \cmidrule(r){1-1}
    \cmidrule(r){2-6}
    \cmidrule(r){7-11}
    WA
    & 0.261
    & 0.332
    & 0.518
    & 0.567
    & 0.611
    & 8.42
    & 30.64
    & 66.41
    & 70.19
    & 75.26 \\    
    $\textup{JUG}^{*}$
     & 0.005  & 0.244  & 0.618 & 0.670 & \textbf{0.726}
     & 0.00  & 33.48 & 67.44  & 79.19 & \textbf{89.17} \\      
    \cmidrule(r){1-1}
    \cmidrule(r){2-6}
    \cmidrule(r){7-11}     
    decoupled &  0.250  & 0.288  & 0.598 & 0.632 & 0.719 
    & 0.00  & 28.21  & 70.24 & 73.46 & 88.45 \\
    joint
    & 0.270
    & 0.348
    & 0.577
    & 0.639
    & 0.658
    & 0.00  & 24.52  & 64.30 & 69.92 & 86.86 \\     
    joint+aug
    & 0.329
    & 0.361
    & 0.589
    & 0.662
    & 0.671
    & 4.21  & 26.33  & 67.43 & 71.19 & 87.10 \\  
    step 1
    & 0.371
    & 0.429
    & 0.570
    & 0.607
    & 0.632
    & 12.19  & 35.89  & 72.90 & 72.01 & 84.73 \\      
    Ours (step 1+2)
    & \textbf{0.401}
    & \textbf{0.458}
    & \textbf{0.644}
    & \textbf{0.672}
    & {0.717}
    & \textbf{16.62}
    & \textbf{42.74}
    & \textbf{75.94}
    & \textbf{80.36}
    & 87.77\\    
    \bottomrule
`  \end{tabular}}  
\caption{\label{tb:comparison} \small Performance for NLG (BLEU-4) and NLU (joint accuracy (\%)) on E2E and Weather datasets with increasing amount of labeled data from 10, 50 labeled instances to 1\%, 5\%, and 100\% of the labeled data ($D_L$). Models that have access to \textbf{unlabeled ground-truth text labels} are marked with *.
We provide results for the NLG and NLU models trained separately using supervised objectives alone (\emph{decoupled}), our semi-supervised joint-learning model (\emph{joint}), \emph{joint} with all unlabeled data (\emph{joint+aug}), and weakly-supervised models (\emph{step 1}). \emph{Step 1+2} denotes the full proposed approach.
}
\end{table*}

\begin{table}[t]
\centering
\resizebox{0.85\columnwidth}{!}{%
\begin{tabular}{c|c|c|c|c|c}
\hline
\multicolumn{1}{l|}{} & $D_{L}$ & $D_W$ & $X_U$ & $Y_{SL}$ & $Y_{WL}$ \\ \hline
$\textup{JUG}$   & \checkmark           & \xmark                & \checkmark
         & \checkmark           & \xmark                      \\ \hline
WA    & \checkmark                 & \xmark                & \xmark          & \xmark            & \xmark                      \\ \hline
decoupled  & \checkmark            & \xmark                & \xmark          & \xmark            & \xmark                      \\ \hline
joint     & \checkmark              & \xmark                & \checkmark         & \xmark            & \xmark                      \\ \hline
joint+aug  & \checkmark             & \xmark                & \checkmark         & \xmark            & \checkmark                     \\ \hline
step 1     & \xmark             & \checkmark               & \checkmark         & \xmark            & \checkmark                     \\ \hline
Ours (step 1+2)   & \checkmark            & \checkmark               & \checkmark         & \xmark            & \checkmark                     \\ \hline
\end{tabular}}
\caption{\label{tb:summary_data} \small Summary of training data used in each model. Sources of data include labeled data ($D_L$), unlabeled MR ($X_U$),  weakly labeled data ($D_W$), 100\% real text ($Y_{SL}$), and weak text labels ($Y_{WL}$). }
\end{table}

\section{Experiment Setting}
\label{sec:exp}
\paragraph{Data.}
We conduct experiments on the  Weather~\cite{balakrishnan2019constrained} and E2E~\cite{novikova2017e2e} datasets. Weather contains 25k instances of tree-structure annotations. E2E is a crowd-sourced dataset containing 50k instances in the restaurant domain. The inputs are dialogue acts consisting of three to eight slot-value pairs. 
\paragraph{Configurations.}
Both NLU and NLG models are implemented in PyTorch~\cite{pytorch} with 2 Bi-LSTM layers and
$200$-dimensional token embeddings and Adam optimizer~\cite{kingma2014adam} with initial learning rate at $0.0002$. 
Batch size is kept at $28$ and we employ beam search with size $3$ for decoding.
The score is averaged over 10 random initialization runs. 
In our implementation, the sequence-to-sequence models are built upon the bi-directional long short-term memory (Bi-LSTM)~\cite{hochreiter1997long}. 
For LSTM cells, both the encoder and decoder have $2$ layers, amounting to 18M parameters for the seq2seq model. 
All models were trained on 1 Nvidia V100 GPU (32GB and CUDA Version 10.2) for 10k steps. 
The average training time for seq2seq model was approximately 1 hour, and roughly 2 hours for the proposed semi-supervised training with 100\% data.
The total number of updates is set to 10k steps for all training and patience is set as $100$ updates. At decoding time, sentences are generated using greedy decoding.

\begin{figure*}[ht]
  \centering
\includegraphics[width=\textwidth]{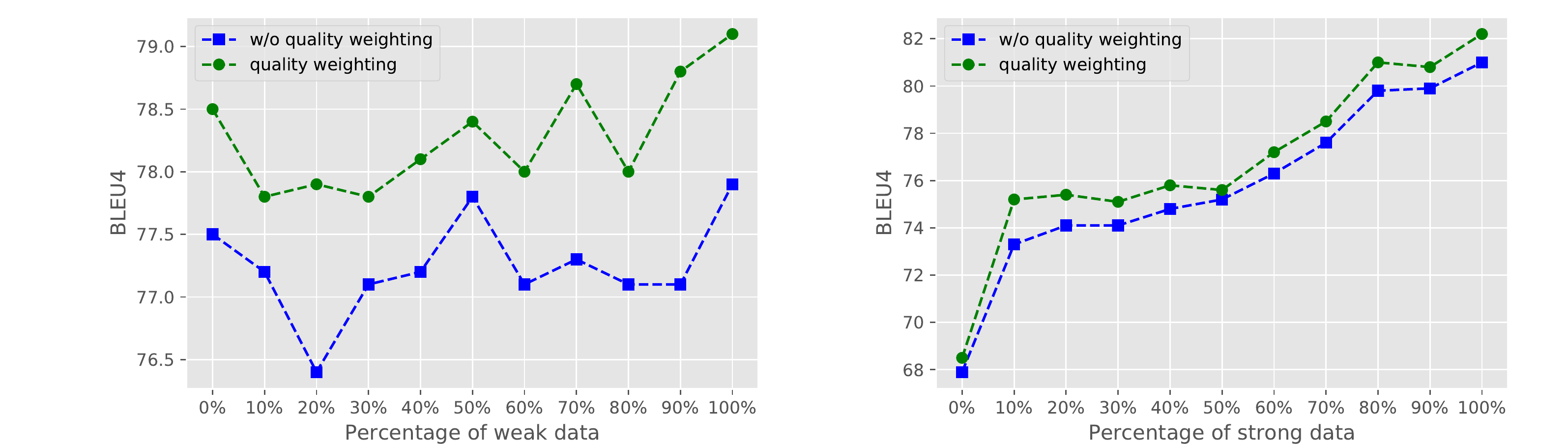}
\caption{\small Model performance (BLEU-4) on $5$\% E2E data with varying percentages of \emph{strong} and \emph{weak} data \emph{with} and \emph{without} DMI-based quality weighting. 
\textbf{Left plot} begins with models trained on labeled data while \textbf{right plot} starts with the weak synthesized dataset instead.
}
\label{fig:two_step}
\end{figure*}

\begin{table}[t]
\centering
\resizebox{0.9\columnwidth}{!}{%
\begin{tabu}{ll}
\tabucline [1pt]{1-2}
\textbf{E2E NLG} & \textbf{BLEU-4} \\
TGEN \cite{duvsek2016sequence} & 0.6593 \\
SLUG \cite{juraska2018deep} & 0.6619 \\
Dual supervised learning \cite{su2019dual} & 0.5716 \\
$\textup{JUG}$~\cite{tseng2020generative} & 0.6855 \\
GPT2-FT~\cite{chen2019few} & 0.6562 \\
WA~\cite{harkous2020have} & 0.6445 \\
Ours (step 1+2) & \textbf{0.7025} \\
\\ \hline
\textbf{Weather NLG} & \textbf{BLEU-4} \\
S2S-CONSTR \cite{balakrishnan2019constrained} & 0.7660 \\
$\textup{JUG}$~\cite{tseng2020generative} & 0.7768 \\
Ours (step 1+2) & \textbf{0.7986} \\
\tabucline [1pt]{1-2}
\end{tabu}%
}
\caption{
\small 
For comparison, we show the performance of previous systems on the datasets following the \textbf{original split}, so the scores are \textbf{not comparable} to Table~\ref{tb:comparison}. }
\label{tab:sota}
\end{table}

\section{Results}

We first compare our model with other baselines on both datasets, then perform a set of ablation studies on the E2E dataset to see the effects of each component. 
Finally, we analyze the strength of the \emph{weak annotator}, and the effect of the quality-weighted weak supervision, before concluding with the analysis of dual mutual information.

In particular, we experiment with various low resource conditions of training set (10 instances, 50 instances, 1\% of all data, 5\% of all data).
To show that our proposed approach is consistently better, we include the scenario with 0-100\% of the data at 10\% interval, to show that performance does not deteriorate as more training samples are added (Figure~\ref{fig:two_step}).
Table~\ref{tb:summary_data} shows the summary of training data used for all models in Table~\ref{tb:comparison}.
We compare our model with 
(1) a fine-tuned GPT2 model (GPT2-FT) that uses a switch mechanism to select between input and GPT2 knowledge~\cite{chen2019few}\footnote{\url{https://github.com/czyssrs/Few-Shot-NLG}}, 
(2) a fine-tuning approach to be used as the weak annotator (WA) that predict text from MR or MR from text, depending on the input format during fine-tuning~\cite{harkous2020have}\footnote{No released source code so we re-implemented it based on paper.}, and 
(3) the semi-supervised model~\footnote{\url{https://github.com/andy194673/Joint-NLU-NLG}} ($\textup{JUG}$) from~\citet{tseng2020generative}.
Note that the specialized encoder in \emph{GPT2-FT} cannot be easily adapted to the tree-structured input in Weather, and so we do not provide its score on the Weather dataset.

In Table~\ref{tb:comparison}, we show that our proposed approach (\emph{step 1+2}) generally performs better than the baselines for both tasks (NLG and NLU) for most selected labeled data sizes.
We show that even with only $10$ labeled instances, our approach (\emph{step 1+2}) is able to yield decent results compared to the baselines.
The difference between models tends to be larger for settings with few training instances, and the advantage of the method diminishes \emph{as the amount of labeled data available for $\textup{JUG}$ increases}, to the point where $\textup{JUG}$ is able to outperform the proposed approach.
Overall, the benefit of the noisy supervisory signal from the weak data is able to boost performance, especially at lower resource conditions.

\begin{table}[t]
\centering
\resizebox{1\columnwidth}{!}{%
\begin{tabu}{lccc|ccc}
\tabucline [1pt]{1-7}
\multirow{2}{*}{Method} & \multicolumn{3}{c}{NLU} & \multicolumn{3}{c}{NLG} \\
 & Miss & Redundant & Wrong & Fluency & Miss & Wrong \\ \hline
decoupled 				& 72 			& 78 			& 87 			& 4.10 		& 69 & 73 \\
$\textup{JUG}$ 			& 65 			& \textbf{72} 	& 75 			& 4.23 		& 64 & 65 \\
Ours (step 1+2) 				& \textbf{54} 	& 77 			& \textbf{68} 	& \textbf{4.50} & \textbf{63} & \textbf{61} \\
\tabucline [1pt]{1-7}
\end{tabu}%
}
\caption{
\small 
Human evaluation on the sampled E2E outputs (100 instances) for models with 1\% training data. Numbers of \emph{missing}, \emph{redundant} and \emph{wrong} predictions on slot-value pairs are reported for NLU; \emph{fluency}, numbers of \emph{missing} or \emph{wrong} generated slot values are listed for NLG. }
\label{tab:error_analysis}
    \vspace{-1.3em}
\end{table}

\begin{figure*}[ht]
  \centering
\includegraphics[width=0.85\textwidth]{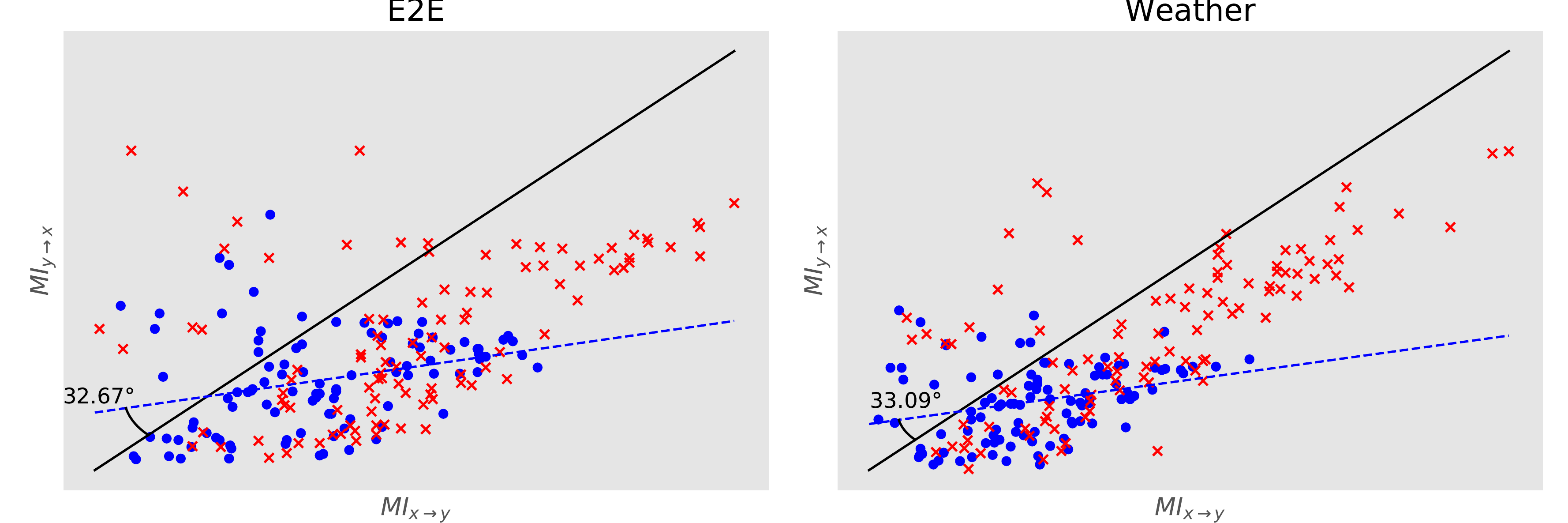}
\caption{\small Visualization of dual mutual information (DMI) on both datasets where
\textbf{\texttimes} markers are $50$ random samples from annotated data and \textbf{$\circ
$} markers are $50$ random samples in the weak dataset.
\textbf{Dotted} lines are trend lines for \textbf{$\circ
$} markers and \textbf{solid black} lines are diagonal reference that correspond to the perfect NLG-NLU balance where both tasks have equal difficulties.
}
\label{fig:dmi}
\end{figure*}

We observe that training with weakly labeled data alone (step 1) is not sufficient, and so strong data is required to provide the supervisory signals necessary (step 2).
Further, the fact that \emph{joint+aug} displays noticeable improvements over \emph{joint} suggests that simply having augmented text helps to improve the encoded latent space as projected by both the NLU and NLG encoders.
This also shows an alternative way to introduce additional in-domain information to both models, even though the NLU model does not benefit directly from additional text.
Importantly, our approach shows that the \emph{weak annotator} is able to bridge the gap as defined by the access to ground-truth text labels in $\textup{JUG}$ -- outperforming it significantly at low resource conditions (10, 50, 1\%, 5\%) with the difference in NLG being as large as 48.7 BLEU points with $10$ instances.
We find that the proposed model also performs well in the high resource (100\% of labeled data) condition, as shown in Table~\ref{tab:sota}. 
Moreover, with 100\% labeled data, our model is still able to produce superior performance over some of the baselines, which shows that weak annotation does capture additional useful patterns that benefit the NLG process.

\section{Analysis}

\paragraph{Error Analysis.}
Since word-level overlapping scores usually correlate rather poorly with human judgements
on fluency and information accuracy~\cite{reiter2009investigation,novikova2017we}, we perform human evaluation on the E2E corpus on 100 sampled generation outputs. 
For each MR-text pair, the annotator is instructed to evaluate the \emph{fluency} (score 1-5, with 5 being most fluent), \emph{miss} (count of MR slots that were missed) and \emph{wrong} (count of included slots not in MR) are presented in Table~\ref{tab:error_analysis}, where fluency scores are averaged over 50 crowdworkers.
We show that with $1\%$ data, both NLU and NLG models yield significantly fewer errors in terms of \emph{misses} and \emph{wrong} facts, while having more fluent outputs.
However, it generates more redundant slot-value pairs which we attribute to the noisy augmentation that ``misguided'' the NLU model.

\begin{table}[t]
\centering
\resizebox{0.95\columnwidth}{!}{%
\begin{tabu}{lc|c}
\tabucline [1pt]{1-4}
\multirow{2}{*}{Method} & \multicolumn{1}{c}{NLG} & \multicolumn{1}{c}{NLU} \\
 & BLEU-4 (Accuracy (\%)) &  Accuracy (\%) (F1) \\ \hline
w/ \emph{DF} & 0.683 (77.69) & 24.71 (0.6443)  \\
with \emph{DF} & 0.703 (79.08) & 27.19 (0.6840) \\
with \emph{WS} &  0.733 (82.65) & 30.23 (0.7028) \\
with \emph{WS+CW} & \textbf{0.754 (86.44)} &  \textbf{34.09 (0.7200)}  \\
\tabucline [1pt]{1-4}
\end{tabu}%
}
\caption{ \small
Ablation study of weak supervision (1\% E2E labeled data $D_L$) including data fidelity (\emph{DF}), the proposed model (step 1+2) with weak supervision (\emph{WS}), and \emph{WS} with quality-weighted weak supervision (\emph{WS+CW}). }
\label{tab:df_ablation}
    \vspace{-1.6em}
\end{table}

   \begin{table*}[ht]
   \centering
    \resizebox{1\linewidth}{!}{
        \begin{tabular}{l|l}
        \toprule
        mr  & [name] Giraffe, [eat type] pub, [area] riverside \\
        synthetic reference & 
        Giraffe is a pub in the riverside of the city just down the street. \\ \\

        
        \toprule
        mr  & [name] Strada, [eat type] restaurant, [food] Italian, [area] city centre, [familyfriendly] no, [near] Avalon\\
        synthetic reference & 
        Strada is an Italian restaurant not for the families! it is near Avalon in the city centre. \\ \\
        
        
        \toprule
        mr  & [name] Cocum, [eat type] restaurant, [food] French, [area] riverside, [familyfriendly] no, [near] Raja Indian Cuisine\\ 
        synthetic reference & Cocum sells French food near Raja Indian Cuisine. \\
 \\ 
        \bottomrule
        \end{tabular}
        \smallskip
    }
    \caption{\small Display of weakly-labeled data samples.
    \label{tb:weak}
    }
    \end{table*}

   \begin{table*}[ht]
   \centering
    \resizebox{0.8\linewidth}{!}{
        \begin{tabular}{ll}
        \toprule
        mr  & [name] Blue Spice, [eat type] coffee shop, [area] city centre \\
        \midrule
        step 1+2 & 
Blue Spice is a coffee shop in the city centre that of the city. \\
        JUG & Blue Spice serves Italian food and is family friendly. \\
        decoupled & 
Blue Spice is an adult Italian coffee shop with high customer rating located in \\ \\

        
        \bottomrule
        \end{tabular}
        \smallskip
    }
    \caption{\small Display of text generations from different models.
    \label{tb:generation}
    }
    \end{table*}

\paragraph{How Strong is the \emph{Weak Annotator}?}
To assess the strength of the weak annotator (WA) itself, we also computed its NLG scores with varying amounts of labeled data (see Table~\ref{tb:comparison}).
We observe that the WA suffers from a performance drop in lower resource conditions (i.e.~0.195 BLEU with 10 labeled instances), when the given training samples are not sufficient for the pretrained model to converge upon a region of in-domain generation. 
However, it yields some quality data when conditioned on a large number of possible MR (i.e. $50$\% data), forming a useful in-domain text set (See Table~\ref{tb:weak}).

\paragraph{Analysis of Weak Supervision.}
In Table~\ref{tab:df_ablation}, we present the results of an ablation study on weak supervision (see \cref{sec: rm}) where the effect of \emph{data fidelity} is stronger on NLU than on NLG, which is due to the nature of the filtering process which removes faulty text labels which influences both $x \rightarrow y$ and $y \rightarrow y$ training directions. 
Next, though weak supervision boosted the model by giving direct supervision in training directions $x \rightarrow y$ and  $y \rightarrow x$, the noisy nature of the augmentation limits its effectiveness.
The model is further improved with the proposed quality-weighted update that takes into account the sample quality and alleviate the influence of poor quality samples. 
Refer to Table~\ref{tb:generation} for output comparison.

\paragraph{Analysis of the Two-Step Training Process.}
As inspired by~\citet{dehghani2018fidelity}, we justify the two-step training process by performing two types of experiments with $5\%$ data (see Figure~\ref{fig:two_step}): 
In the first experiment, we use all the available strong data but consider different ratios of the entire weak dataset -- as used in our 2-step approach. 
In the second, we fix the amount of weak data and provide the model with varying amounts of strong data.
The results show that the \emph{student} models are generally better off by having the \emph{teacher}'s supervision. 
Further, pretraining on weak data prior to fine-tuning on strong data appears to be the better approach and this motivates the reasoning behind our two-step approach.

\paragraph{Analysis of the Dual Mutual Information.}
Figure~\ref{fig:dmi} depicts DMI with the visualization of $MI_{x \rightarrow y}$ as x-axis and $MI_{y \rightarrow x}$ as y-axis, in which $100$ randomly sampled noisy and ground-truth samples are plotted for both datasets.
On the plot, the diagonal reference represents the scenario in which NLG and NLU inference are equally difficult, and we see that annotated data cluster more around the diagonal reference.
This means that expert-labeled samples' DMI scores tend to be smaller, where NLU and NLG inference for these samples carry similar levels of difficulty.
Importantly, since DMI scores are normalized over both clean and noisy samples, \emph{the proximity of data to the trendlines can then be used to estimate the sample quality} -- clean data are closer as compared to the noisy samples. 
Thus clean data will have smaller normalized scores, higher $c(\cdot)$, and a larger update step.
This further supports the use of the proposed sample quality-based updates on the parameters.

\section{Conclusion and Future Work}
In this paper, we show the efficacy of the framework where data is automatically labeled and both NLU and NLG models learn with quality-weighted weak supervision so as to account for the individual data quality.
Most importantly, we show that not only is the two-step training process useful in improving the model, it yields decent quality text.
This work serves as a starting point for weakly-supervised learning in natural language generation, especially for topics related to instance-based weighting approaches.

For future work, we hope to extend on the framework and propose ways with which it can be incorporated into existing text annotation systems.
Specifically, we would like to see its effectiveness in human-in-the-loop settings~\cite{crook2018conversational,hong2019improving,de2018generating,zhuang2017neobility,chang2020unsupervised,wiehr2020safe,shen2020neural,chang2020dart,su2020moviechats,chang2021does,chang2021neural} where the quality estimation metrics can take signals from human feedbacks into account. 


\section*{Acknowledgements}
This research was funded in part by the German Research Foundation (DFG) as part of SFB 248 ``Foundations of Perspicuous Software Systems''. We sincerely thank the anonymous reviewers for their insightful comments that helped us to improve this paper.

\bibliography{anthology,eacl2021}
\bibliographystyle{acl_natbib}

\end{document}